\documentclass[journal]{IEEEtran}
\usepackage{amsmath,amssymb,graphicx,booktabs,xcolor,url,multirow,rotating}
\usepackage[hidelinks]{hyperref}

\begin{document}

\title{Evaluating Generative Time-Series Models\\ on Data with Point Masses}

\author{Jian~Xu$^{1,2}$%
\thanks{$^{1}$RIKEN iTHEMS, Wako, Japan.
$^{2}$RIKEN Center for Advanced Intelligence Project (AIP), Tokyo, Japan.
E-mail: \texttt{jian.xu@riken.jp}}%
\thanks{Preprint, \today. This work has not been peer reviewed.}}

\maketitle

\begin{abstract}
Many of the series that generative time-series models are benchmarked on place a
large probability mass on a single value --- it does not rain, no ride is
requested, no part is ordered. We report what happens when such data is
evaluated carefully. First, the standard rolling-origin protocol can score a
model on a window whose atom structure bears no resemblance to the dataset: on
one benchmark the dataset is $42\%$ zeros and the evaluation windows are $13\%$,
on another $47\%$ against $5\%$. This is not a cosmetic problem --- it reversed
one of our own conclusions, turning the strongest occurrence model in our study
into what looked like a cautionary tale. Second, we give a control in which CRPS
is invariant \emph{by construction} while the temporal coupling is destroyed,
which measures exactly how much that coupling contributes to a chosen statistic.
Third, benchmarking seven models on a matched protocol over five seeds, an
autoregressive hurdle beats a conditional flow on five of six datasets, by up to
a factor of $153$, while the flow's own occurrence statistics vary by up to
$62\%$ across training seeds and every baseline is deterministic. Finally, the
model ordering is not the same under five different occurrence statistics, and
the two that do not share a construction agree with each other least.
\end{abstract}


\IEEEpeerreviewmaketitle

\section{Introduction}
\label{sec:intro}

A large share of the time series that generative models are trained and
evaluated on take one particular value with positive probability. It does not
rain on most days; no ride is requested in most hours; no spare part is ordered
in most months; a solar panel produces nothing at night. In several of the
datasets that the recent generative time-series literature reports on, between
a quarter and nine tenths of all observations are exactly zero
(Table~\ref{tab:atoms}). The data law is a mixture: an atom at zero, and a
continuous distribution over the positive part.

The models are not built that way. Flow matching with data-dependent priors
\cite{kollovieh2025flow}, residual-expert flows \cite{zhang2026prismflow},
stochastic flow matching \cite{panjing2025timeflow}, VQ-latent transport
\cite{li2026sdflow} and their diffusion predecessors
\cite{wang2025non,ye2025non} all transport an absolutely continuous source
distribution with a learned map. Under the regularity conditions ODE-based flow
models are usually analysed with --- a velocity field Lipschitz in the state and
integrable in time, so that the time-$1$ flow map is a diffeomorphism --- the
pushforward of an absolutely continuous law is again absolutely continuous, and
\begin{equation}
  \Pr\nolimits_{\text{model}}(Y_t = c) = 0 \quad \text{for every } c \in \mathbb{R}.
  \label{eq:nomass}
\end{equation}
Continuity of the map alone would not give this --- a continuous map can collapse
a set to a point --- but invertibility does, and the model we study is an
ODE-based flow \cite{kollovieh2025flow} integrated with a fixed-step solver, so
its sampler realises an invertible map by construction. Diffusion samplers reach
the same place by a different route: the exact reverse-time limit can converge to
a singular law, but the truncation at $t = \epsilon > 0$ and the finite step size
used in practice leave the output continuous.
A model is therefore being asked to reproduce, on more than half of the
observations in some of these datasets, a feature it cannot represent. The
scores used to judge it are marginal --- CRPS per timestep, or a distributional
distance computed on pooled windows --- and are not obviously able to detect the
difference between a genuine atom and a thin spike of continuous mass near zero.

This paper began as a test of the hypothesis those two facts suggest: that
continuous-state models are systematically handicapped on data with a point
mass, and that an explicit occurrence process would fix it. Neither part
survived contact with the experiments. Across six datasets the flow is ahead of every explicit
occurrence model on one dataset only, and an autoregressive occurrence hurdle
--- a logistic classifier rolled out one step at a time --- beats it on the
other five, on \texttt{rideshare} by a factor of $153$. The empirical gap is real; what is
not supported is that a continuous state is its cause, or that closing it
requires a new generative model.

What the experiments did produce is a set of problems with the measurement, and
they are the contribution of this paper. We distinguish two kinds of statement
throughout. The \emph{evaluation-level} findings --- what a protocol measures,
what a marginal score can support --- follow from the definitions and apply to
any generative model scored this way. The \emph{model-level} findings are
empirical claims about one family of conditional flows on six datasets; we do
not read them as properties of continuous-state generative modelling in
general.

\begin{itemize}\itemsep2pt
  \item \textbf{The evaluation window need not resemble the dataset.} The
    standard rolling-origin protocol scores a model on the final $H$ steps of
    each series. Where the atom process is non-stationary, those steps can be
    unrepresentative by a wide margin: \texttt{covid\_deaths} is $42\%$ zeros as
    a dataset and $13\%$ in its evaluation windows, \texttt{rideshare} $47\%$
    against $5\%$. We show below that this inverted one of our own findings
    (Sec.~\ref{sec:protocol}).
  \item \textbf{A control under which CRPS cannot move.} Permuting the sample
    index independently at each horizon step leaves every per-step predictive
    marginal --- and therefore CRPS --- unchanged by construction, while
    destroying the coupling that carries the occurrence statistics. It measures
    exactly how much the model's trajectory coupling contributes to a chosen
    statistic (Sec.~\ref{sec:control}).
  \item \textbf{A benchmark over five seeds, and the model that wins it.} An
    autoregressive occurrence hurdle beats the flow on five of six datasets, by
    up to $153\times$; the flow's own occurrence statistics move by up to $62\%$
    across training seeds while every baseline is deterministic; and the model
    ordering changes with the choice of occurrence statistic
    (Sec.~\ref{sec:bench}, Sec.~\ref{sec:diag}).
\end{itemize}

Substituting the conventional semi-Markov occurrence process into the flow,
however, makes the relevant statistic worse on five of six datasets: the
occurrence model that works here is the autoregressive one, and which explicit
process is substituted matters more than whether one is.

\section{Related work}
\label{sec:related}

\subsection{Generative models for time series}
Adversarial models \cite{yoon2019time,jeha2022psa} were displaced by
diffusion --- autoregressive over the horizon \cite{rasul2021autoregressive},
conditioned on the observed part \cite{tashiro2021csdi}, or decomposed into
interpretable components \cite{yuan2024diffusion} --- and more recently by
flow matching \cite{lipman2023flow}, which is now the default for
probabilistic forecasting and unconditional generation of time series. Work in this line has concentrated on the transport: TSFlow
\cite{kollovieh2025flow} replaces the isotropic source with a Gaussian-process
prior matched to the temporal structure of the data; PrismFlow
\cite{zhang2026prismflow} attributes the spectral contraction of a single global
velocity estimator to incompatible conditional velocities and adds
Koopman-inspired residual experts; TimeFlow \cite{panjing2025timeflow} restores with
an SDE the randomness an ODE formulation discards; SDFlow \cite{li2026sdflow}
moves the transport into a frozen VQ latent space. On the diffusion side the
forward operator has been adapted to the spectral profile of time series
\cite{wang2025non} and to time-varying uncertainty \cite{ye2025non}. All of
these are continuous-state in the sense of Eq.~\ref{eq:nomass}, all report on
overlapping subsets of the datasets in Table~\ref{tab:atoms}, and none of them
discusses the atom.

\subsection{Intermittent series, occurrence processes and censoring}
Outside generative modelling the point mass defines a mature subfield, and the
baselines of Sec.~\ref{sec:bench} are taken directly from it. Croston's method
\cite{croston1972forecasting} splits demand occurrence from demand size, with a standard
bias correction \cite{syntetos2005accuracy}, neural variants
\cite{kourentzes2013intermittent} and a probabilistic treatment via deep renewal
processes \cite{turkmen2021forecasting}; even the accuracy measures used there need care on
intermittent data \cite{hyndman2006another}. In hydrology, chain-dependent
weather generators have paired a Markov chain for wet/dry occurrence with a
distribution for the amounts since Katz \cite{katz1977precipitation} and Richardson
\cite{richardson1981stochastic}, with Stern and Coe \cite{stern1984model} fitting
non-stationary occurrence chains to reproduce dry-spell lengths --- the
statistic we use throughout --- and Wilks \cite{wilks1999simultaneous} extending it
to multiple sites. Where the atom comes from the observation process instead ---
detection limits, saturation, clipping --- the setting is censored regression
\cite{tobin1958estimation,amemiya1984tobit}. What none of this provides is joint
trajectory generation: it works with marginal or low-order conditional forecasts
and is absent from the benchmarks the generative models report on. Running the
two families against each other on the same windows is one contribution of this
paper.

\subsection{Evaluating generative time series}
Reported comparisons rest on CRPS --- a strictly proper score, but a marginal
one \cite{gneiting2007strictly} --- and, for unconditional generation, on a
discriminative score and Context-FID \cite{jeha2022psa}, a Fr\'echet distance
in the representation space of a contrastive encoder \cite{yue2022ts2vec}. Both
are marginal or pooled. Scores that are sensitive to dependence exist --- the
energy score \cite{gneiting2007strictly} and, more sharply for correlation
structure, the variogram score \cite{scheuerer2015variogram} --- and we report
both, but neither isolates the contribution of the coupling from that of the
marginals. The control of Sec.~\ref{sec:control} does, and it is the inverse of
a familiar operation: the Schaake shuffle \cite{clark2004schaake} and ensemble
copula coupling \cite{schefzik2013uncertainty,schefzik2016similarity} \emph{impose} a
rank dependence on samples that were postprocessed marginally, whereas we
\emph{remove} the dependence a model produced while holding those marginals
fixed. Used forward it is a correction; used backward it is a measurement.

\section{Atoms in time-series benchmarks}
\label{sec:atoms}

Table~\ref{tab:atoms} reports, for datasets drawn from the GluonTS and Monash
repositories \cite{godahewa2021monash}, the fraction of exact zeros and two
descriptors of the atom process: the calendar-unexplained occurrence entropy
\begin{equation}
  s = H(Z \mid \text{calendar}) \,/\, H(Z), \qquad Z_t = \mathbf{1}\{Y_t = 0\},
  \label{eq:sent}
\end{equation}
estimated with a cross-fitted logistic model on calendar harmonics, and the
over-dispersion of the zero-run law relative to a geometric distribution,
\begin{equation}
  \phi = \mathrm{Var}(L) \,/\, \bigl(\mathbb{E}[L]^2 - \mathbb{E}[L]\bigr),
  \qquad \phi = 1 \text{ for geometric.}
  \label{eq:phi}
\end{equation}
Both are computed from the raw data, with no model and no access to the test
split. Neither turns out to predict which datasets a flow struggles on
(Section~\ref{sec:bench}); we report them because they are the two natural
candidates and because ruling them out is itself informative.

\begin{table}[t]
\caption{Atom mass and two data-only descriptors. $s$ is the share of
occurrence uncertainty the calendar cannot explain (Eq.~\ref{eq:sent}); $\phi$
is the over-dispersion of the zero-run law (Eq.~\ref{eq:phi}). Neither predicts
which datasets the flow struggles on (Sec.~\ref{sec:bench}).}
\label{tab:atoms}
\centering
\footnotesize
\setlength{\tabcolsep}{4pt}
\begin{tabular}{lccc}
\toprule
Dataset & $\Pr(Y{=}0)$ & $s$ & $\phi$ \\
\midrule
\texttt{car\_parts}          & 0.94 & 1.00 & 0.72 \\
\texttt{weather}$^\dagger$   & 0.62 & 1.00 & 1.88 \\
\texttt{temp\_rain}$^\dagger$& 0.57 & 0.66 & 7.93 \\
\texttt{solar}               & 0.50 & 0.15 & 0.02 \\
\texttt{rideshare}           & 0.47 & 0.93 & 1.47 \\
\texttt{covid}$^\ddagger$    & 0.42 & 0.65 & 0.43 \\
\texttt{uber}                & 0.31 & 0.98 & 30.7 \\
\midrule
\texttt{electricity}         & 0.01 & --- & --- \\
\texttt{traffic}             & 0.00 & --- & --- \\
\bottomrule
\end{tabular}

\smallskip
{\scriptsize $^\dagger$mixed: these interleave rain series with temperature and
pressure series that have no atom ($74\%$ of \texttt{weather}'s series have
$\Pr(Y{=}0)<0.01$); we use the $\Pr(Y{=}0)\ge0.3$ subset throughout.
$^\ddagger$excluded from Sec.~\ref{sec:bench}, see Sec.~\ref{sec:protocol}.}
\end{table}

\section{Two evaluation problems}

\subsection{The evaluation window need not resemble the dataset}
\label{sec:protocol}

\begin{figure}[t]
\centering
\includegraphics[width=0.8\linewidth]{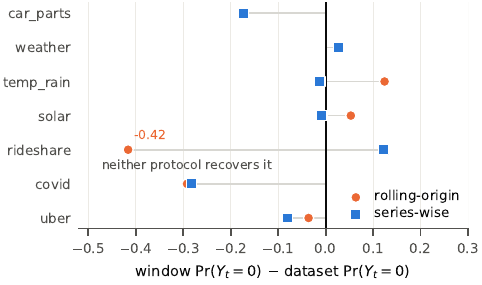}
\caption{How far the evaluation window departs from the dataset it is drawn
from. Zero is agreement. Under the rolling-origin protocol
\texttt{rideshare}'s windows are $42$ points short of the dataset's zero rate,
so the protocol is not testing the atom behaviour it appears to. A series-wise
split fixes that case but not \texttt{covid}, whose zeros belong to the
pre-outbreak period, and introduces a departure of its own on
\texttt{car\_parts}.}
\label{fig:protocol}
\end{figure}

The standard protocol forecasts the final $H$ steps of each series. When the
atom process is non-stationary, those windows can be unrepresentative of the
dataset by a wide margin. On \texttt{covid\_deaths} the dataset is $42.3\%$
zeros while the evaluation windows are $13.1\%$; on \texttt{rideshare} it is
$46.9\%$ against $5.3\%$ (Fig.~\ref{fig:protocol}). Any statement about atom behaviour measured under
this protocol on these datasets is therefore measured on a regime the dataset
barely contains.

A series-wise split --- hold out whole series, draw windows across their full
time axis --- removes the problem where the atom process is stationary in time:
on \texttt{rideshare} the evaluation-window zero rate moves from $5.3\%$ to
$59.1\%$. It does not rescue \texttt{covid\_deaths} ($14.1\%$), whose zeros are
an artefact of the pre-outbreak period rather than a recurring process; we
exclude it below and state the reason rather than reporting a number that
cannot mean what it appears to.

\subsection{A control under which CRPS cannot move}
\label{sec:control}

CRPS is a function of the per-timestep predictive marginals alone, so it is
often argued that it cannot see temporal structure. We make this measurable.
Given samples $\hat{Y}^{(1:K)}_{1:H}$ for a window, permute the sample index
independently at each step $h$:
\begin{equation}
  \tilde{Y}^{(k)}_h = \hat{Y}^{(\pi_h(k))}_h , \qquad
  \pi_h \sim \mathrm{Unif}(\mathfrak{S}_K) \text{ i.i.d.}
  \label{eq:decorr}
\end{equation}
The multiset of values proposed at each $(\text{window}, h)$ is unchanged, so
every per-step marginal --- and hence CRPS --- is invariant by construction,
while the coupling that carries the run-length law is destroyed. We compute the
empirical CRPS exactly, evaluating $\mathbb{E}|X-X'|$ with the closed form for
the mean absolute pairwise difference of a sorted sample rather than from a
random pairing; the measured difference under Eq.~\ref{eq:decorr} is then $0$ to
floating-point precision on every dataset. (A pairing-based estimator of the
second term is \emph{not} invariant even though CRPS is, and reports a spurious
difference of order $10^{-2}$ --- a detail worth stating because it is easy to
mistake for a real effect.)

\begin{figure}[t]
\centering
\includegraphics[width=0.8\linewidth]{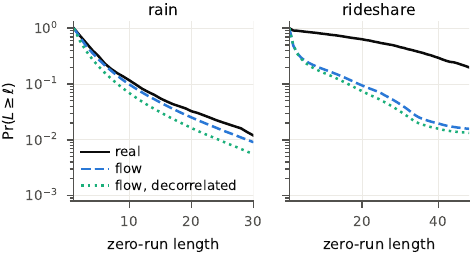}
\caption{Zero-run-length CCDFs. On \texttt{rain} the flow tracks the data and
its decorrelated version does not, so the learned coupling is doing work. On
\texttt{rideshare} the flow is far below the data at every length --- its dry
spells average $5.8$ steps against $26.7$ --- and is indistinguishable from its
own decorrelated samples, so the learned coupling contributes nothing to the
zero-run statistic. CRPS is identical for the dashed and dotted curves by
construction.}
\label{fig:ccdf}
\end{figure}

What the control licenses is narrower than ``the part of the temporal structure
CRPS cannot see'': it is an exact within-model measurement of the contribution
that the model's trajectory coupling makes to a chosen statistic, with the
predictive marginals held fixed. It does not enumerate everything a marginal
score misses, and a model whose zero-run error is unchanged by decorrelation may
still have learned autocorrelation, seasonality, or magnitude--occurrence
coupling that this statistic does not register. Within that scope it gives a
reference point a model must beat: if decorrelating its own samples does not
worsen a statistic, the coupling it learned contributes nothing to that
statistic. Fig.~\ref{fig:ccdf} shows it for two datasets and the last row of
Table~\ref{tab:bench} for all six. On \texttt{uber} and \texttt{car\_parts} the
decorrelated flow is \emph{better} than the flow, and on \texttt{rideshare} the
two are within $12\%$; the learned coupling pays for itself only on
\texttt{rain}, \texttt{weather} and \texttt{solar}.

\section{Benchmark and a negative result}
\label{sec:bench}

We compare, on identical windows and with an identical threshold protocol
(threshold selected on a validation half, test-optimal threshold reported as a
diagnostic upper bound): a conditional flow \cite{kollovieh2025flow}; global
and per-series Markov occurrence chains; a negative-binomial semi-Markov
occurrence process; autoregressive and non-autoregressive hurdle models with a
learned occurrence classifier; and a hybrid that keeps the flow's own positive
magnitudes and replaces only its zero pattern with the semi-Markov process.

\begin{table}[t]
\caption{Zero-run-length $W_1$ (lower is better) under the series-wise
protocol, identical windows and thresholds for every row. The flow is
mean\,$\pm$\,s.d.\ over five training seeds; the occurrence baselines carry no
training stochasticity (closed-form transitions, fixed \texttt{random\_state})
and are therefore reported once. \emph{Ctrl} is the flow's own decorrelated
version (Eq.~\ref{eq:decorr}) --- what a model must beat for its trajectory
coupling to contribute positively to this statistic.}
\label{tab:bench}
\centering
\footnotesize
\setlength{\tabcolsep}{2pt}
\renewcommand{\arraystretch}{1.05}
\begin{tabular}{@{}l@{\hskip 3pt}rrrrrr@{}}
\toprule
& rain & solar & uber & ridesh & weath & c.parts \\
\midrule
Flow & \begin{tabular}{@{}r@{}}.33\\[-2pt]{\tiny$\pm$.11}\end{tabular}
     & \begin{tabular}{@{}r@{}}\textbf{.71}\\[-2pt]{\tiny$\pm$.44}\end{tabular}
     & \begin{tabular}{@{}r@{}}.24\\[-2pt]{\tiny$\pm$.09}\end{tabular}
     & \begin{tabular}{@{}r@{}}19.8\\[-2pt]{\tiny$\pm$1.4}\end{tabular}
     & \begin{tabular}{@{}r@{}}.52\\[-2pt]{\tiny$\pm$.23}\end{tabular}
     & \begin{tabular}{@{}r@{}}.37\\[-2pt]{\tiny$\pm$.03}\end{tabular} \\
Markov, global & .84 & 2.60 & 1.38 & 3.69 & .73 & .57 \\
Markov, series & .22 & 2.63 & .15 & 4.79 & .19 & .45 \\
NB semi-Markov & .90 & 2.58 & .46 & 3.36 & .62 & 1.04 \\
Hurdle, non-AR & 1.06 & 5.01 & .21 & 9.50 & .90 & .23 \\
\textbf{Hurdle, AR} & \textbf{.31} & 1.23 & \textbf{.08} & \textbf{.13} & \textbf{.10} & \textbf{.05} \\
Hybrid & .92 & 2.61 & .46 & 3.38 & .59 & 1.05 \\
\midrule
\emph{Ctrl}: decorrelated & 1.12 & .93 & .35 & 21.0 & 1.37 & .28 \\
\midrule
\multicolumn{7}{@{}l}{\emph{CRPS on the same samples}}\\
Flow & \textbf{3.03} & \textbf{5.50} & \textbf{3.10} & 3.11 & 1.81 & .43 \\
Markov, global & 4.48 & 25.4 & 9.83 & 2.74 & 1.79 & .38 \\
Markov, series & 4.23 & 25.2 & 8.61 & 2.86 & 1.81 & .40 \\
NB semi-Markov & 4.41 & 25.6 & 10.2 & 2.78 & 1.80 & .41 \\
Hurdle, non-AR & 3.96 & 34.9 & 8.70 & 2.58 & 1.80 & \textbf{.36} \\
Hurdle, AR & 4.01 & 20.4 & 8.79 & \textbf{2.27} & \textbf{1.76} & \textbf{.36} \\
Hybrid & 3.87 & 24.3 & 7.00 & 3.25 & 1.78 & .44 \\
\bottomrule
\end{tabular}
\end{table}

Three observations follow from Table~\ref{tab:bench}.

\textbf{An autoregressive hurdle dominates.} A learned occurrence classifier
rolled out one step at a time is the best model on five of the six datasets, and
not narrowly: on \texttt{rideshare} it reaches $0.13$ against the flow's $19.8$,
a factor of $153$. It is also the best or second-best model under the spectral,
spell-quantile and survival statistics of Sec.~\ref{sec:diag}. The lower panel
of Table~\ref{tab:bench} shows why this is not simply a better model: on
\texttt{rain}, \texttt{solar} and \texttt{uber} the flow has the best CRPS of
the seven while being beaten or matched on the occurrence statistics, and on
\texttt{solar} it is ahead on both. The two families of number do not order the
models the same way, which is the situation a practitioner is actually in. We report this arm prominently
because we initially omitted it: under the rolling-origin protocol its rollout
drifts and it looks like a cautionary tale about exposure bias, and we had
written it up as one. Under the series-wise protocol --- where the evaluation
windows actually contain the atom --- it is the strongest occurrence model we
tested. The earlier reading was an artefact of the evaluation windows of
Sec.~\ref{sec:protocol}, which is a second, uncomfortable instance of this
paper's own first finding.

\textbf{The flow's atom statistics are seed-unstable; the baselines' are not.}
Over five training seeds evaluated on identical held-out series and windows, the
flow's run-length error has a relative spread of $7\%$ to $62\%$; on
\texttt{solar} the seeds range over a factor of four. Every occurrence baseline
here is deterministic: the transition probabilities are closed-form and the one
learned component has a fixed seed. A single-run number for a flow on these
statistics is not reportable at the margins these comparisons are decided by,
and the papers we drew the model from report one.

\textbf{The obvious fix is not one, but the obvious fix was the wrong one.} The
hybrid keeps the flow's magnitudes and substitutes an explicit occurrence
process, isolating that component. With a semi-Markov occurrence process it
improves on the flow on \texttt{rideshare} alone and is worse on the other five.
That is the result we would have reported had we stopped there. Table~\ref{tab:bench}
shows why it is the wrong conclusion to draw: the occurrence process that works
is not the semi-Markov one but the autoregressive classifier, and a hybrid built
on that component is the natural next experiment rather than a closed question.

Comparing each flow against its own decorrelated version isolates the value of
the coupling it learned, with the marginals held fixed. Averaged over seeds, the
coupling pays for itself on \texttt{rain} ($3.4\times$), \texttt{weather}
($2.6\times$), \texttt{uber} ($1.5\times$) and \texttt{solar} ($1.3\times$), is
neutral on \texttt{rideshare} ($1.06\times$), and is actively harmful on
\texttt{car\_parts} ($0.82\times$); the occurrence ACF gives the same ordering.
An earlier three-seed version of this table put \texttt{uber} and
\texttt{car\_parts} both on the harmful side --- with five seeds only
\texttt{car\_parts} remains there, which is itself an illustration of the
previous paragraph.

\subsection{Does the conclusion survive a change of statistic?}
\label{sec:diag}

Everything above is measured with the Wasserstein distance between zero-run
length laws. That is one hand-built statistic, and a reader is entitled to ask
whether the picture is an artefact of it. We therefore repeat the comparison
under four more, in two groups. Spell quantiles (mean relative error at the
$50$th, $90$th and $99$th percentile) and the sup-norm between run-length
survival functions are further summaries of the \emph{same} object, so they
corroborate the $W_1$ rather than test it independently: the $W_1$ weights how
far misplaced mass moved, the sup-norm weights how much mass is misplaced, and a
model can do well on one and badly on the other. The occurrence autocorrelation
and the normalised occurrence periodogram are independent of the run-length
construction --- they read the same binary process through a second-order and a
spectral lens, and are defined for windows containing no completed spell. We
also report the two transition rates of the occurrence chain.

\begin{table}[t]
\caption{Mean rank across the six datasets under each occurrence statistic
(1 = best of seven models), five seeds. The ordering is not the same under every
statistic, and the two that do not share the run-length construction
(ACF, spectrum) agree with each other less than either agrees with $W_1$.}
\label{tab:diag}
\centering
\footnotesize
\setlength{\tabcolsep}{4pt}
\begin{tabular}{@{}lccccc@{}}
\toprule
& \multicolumn{3}{c}{run-length family} & \multicolumn{2}{c}{independent} \\
\cmidrule(lr){2-4}\cmidrule(lr){5-6}
Model & $W_1$ & spellQ & survS & ACF & spec \\
\midrule
Flow                & 3.2 & 2.7 & 2.8 & \textbf{2.0} & 3.0 \\
Markov, global      & 5.3 & 4.2 & 4.5 & 5.8 & 4.7 \\
Markov, series      & 3.3 & 3.0 & 3.0 & 4.7 & 2.7 \\
NB semi-Markov      & 4.8 & 4.3 & 4.3 & 4.7 & 4.7 \\
Hurdle, non-AR      & 4.5 & 4.5 & 4.3 & 3.5 & 4.8 \\
\textbf{Hurdle, AR} & \textbf{1.3} & \textbf{1.2} & \textbf{1.2} & 2.2 & \textbf{1.3} \\
Hybrid              & 5.5 & 5.2 & 5.0 & 5.2 & 5.0 \\
\bottomrule
\end{tabular}

\smallskip
{\scriptsize Rank correlation between orderings, averaged over datasets:
$W_1$--survS $+0.92$, $W_1$--spellQ $+0.84$, $W_1$--spec $+0.79$,
$W_1$--ACF $+0.55$, ACF--spec $+0.45$.}
\end{table}

\section{Conclusion}

A conditional flow is beaten on five of six atom-bearing datasets --- on
\texttt{rideshare} by a factor of $153$ --- by an autoregressive occurrence
hurdle with no generative machinery to speak of, yet it has the best CRPS of the
seven models on three of those same datasets, so which model is better depends
on which family of number one reads and on which occurrence statistic within the
second family. Three things follow that do not depend on the hypothesis we
started from or on the model family we tested: report the atom rate of the
\emph{evaluation windows}, since the standard protocol can differ from the
dataset by forty points and inverted the ranking of the best baseline in this
study before we caught it; compare a model against its own decorrelated samples
before crediting it with temporal structure, which costs one permutation, holds
CRPS fixed by construction, and reveals cases where the learned coupling
contributes little or negatively; and report a spread, since this flow's
occurrence statistics move by up to $62\%$ across training seeds while every
occurrence baseline here is deterministic. 

\bibliographystyle{IEEEtran}
\bibliography{main}

\end{document}